\documentclass{article}

\usepackage[utf8]{inputenc}  % general document formatting
\usepackage{amsmath,amssymb,amsfonts,amsthm}  % math symbols
\usepackage{geometry} % page size and margins
\usepackage{hyperref}  % use links
\hypersetup{colorlinks=true}
\usepackage{booktabs}  % display a CSV
\usepackage{pifont}  % for cross and checkmarks
\usepackage{graphicx}  % use graphics
\graphicspath{ {./figures/} }
\usepackage{subcaption}  % caption for multi-figures
\usepackage{tikz}
\usetikzlibrary{bayesnet}
\usetikzlibrary{arrows}
\usepackage[ruled, vlined]{algorithm2e}  % display algorithms
\usepackage{datetime}  % define date
\newdate{date}{15}{02}{2021}

\newcommand{\modelfull}{Composable Generative Model}
\newcommand{\model}{\emph{CGM}}

\title{Composable Generative Models}
\author{
    Johan Leduc \\ \href{mailto:jl@sarus.tech}{jl@sarus.tech}
    \and
    Nicolas Grislain \\ \href{mailto:ng@sarus.tech}{ng@sarus.tech}
}
\date{\displaydate{date}}

\begin{document}

\maketitle

\begin{abstract}
Generative modeling has recently seen many exciting developments with the
advent of deep generative architectures such as Variational Auto-Encoders (VAE)
or Generative Adversarial Networks (GAN). The ability to draw synthetic
\emph{i.i.d.} observations with the same joint probability distribution as a
given dataset has a wide range of applications including representation
learning, compression or imputation. It appears that it also has
many applications in privacy preserving data analysis, especially when used in
conjunction with \emph{differential privacy} techniques.

This paper focuses on synthetic data generation models with privacy preserving
applications in mind. It introduces a novel architecture, the
\emph{\modelfull{}} (\model) that is \emph{state-of-the-art} in tabular data
generation. Any conditional generative model can be used as a sub-component of
the \model, including \model s themselves, allowing the generation of
numerical, categorical data as well as images, text, or time series.

The \model{} has been evaluated on 13 datasets (6 standard datasets and 7 simulated)
and compared to 14 recent generative models. \emph{It beats the state of the art in tabular data
generation by a significant margin}.
\end{abstract}

\begin{figure}
    \centering
    \begin{subfigure}{0.36\textwidth}
        \includegraphics[width=\linewidth]{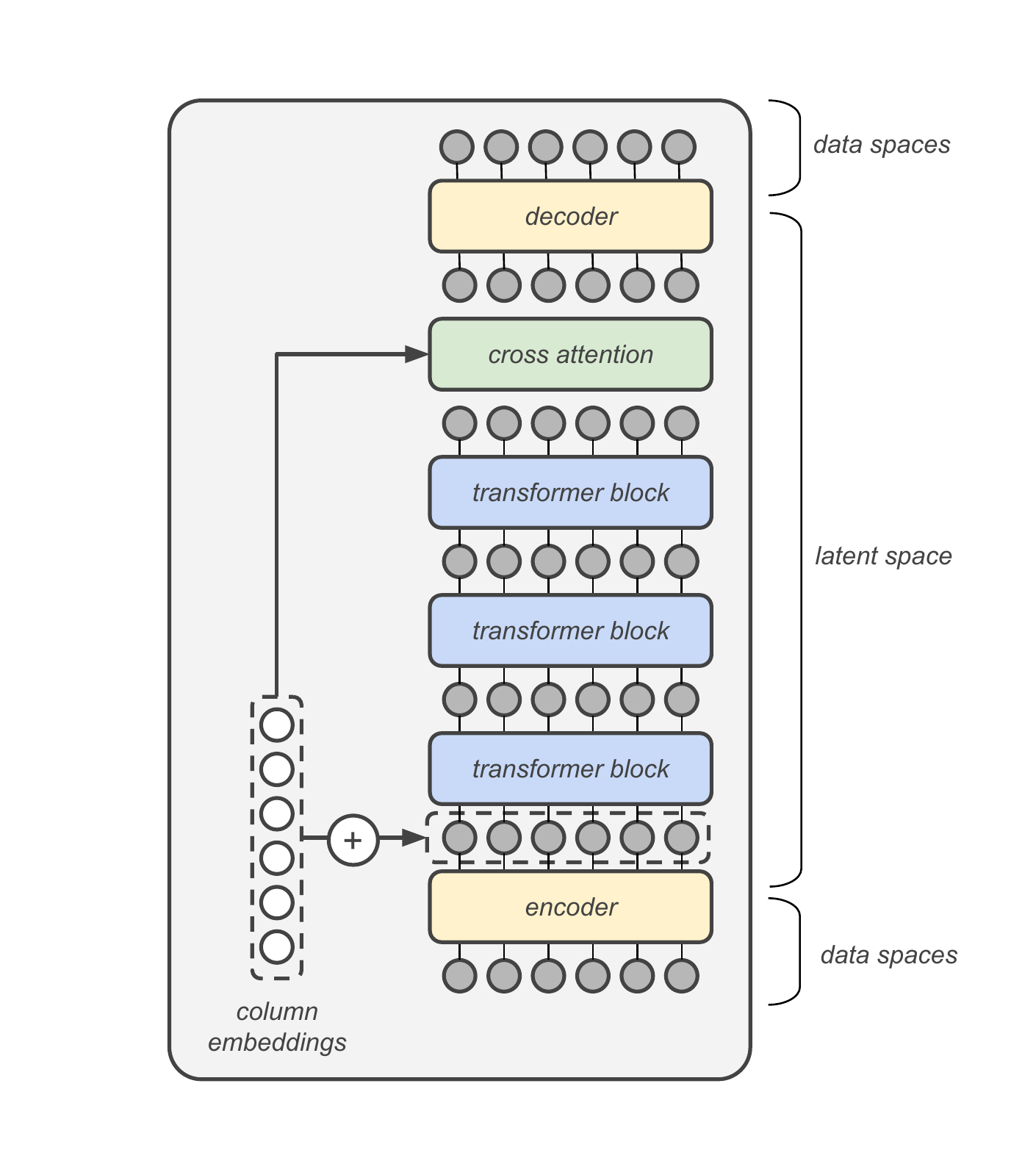}
        \caption{Model overview} \label{fig:1a}
    \end{subfigure}
    \hspace*{\fill}
    \begin{subfigure}{0.21\textwidth}
        \includegraphics[width=\linewidth]{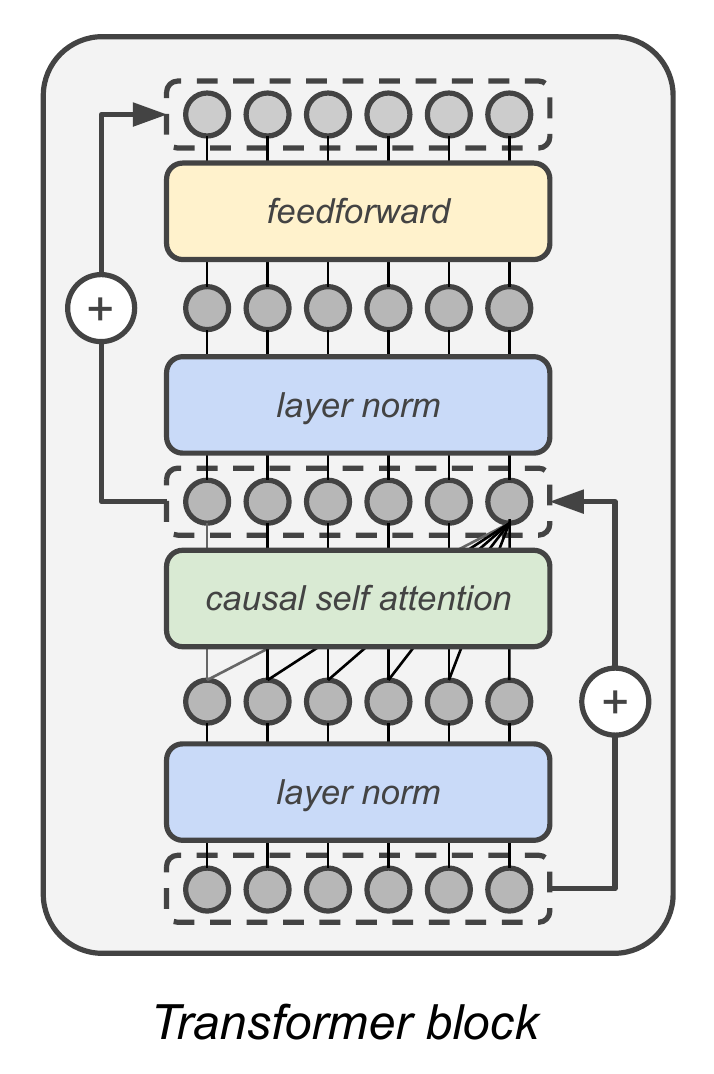}
        \caption{Transformer block detail} \label{fig:1b}
    \end{subfigure}
    \hspace*{\fill}
    \begin{subfigure}{0.31\textwidth}
        \includegraphics[width=\linewidth]{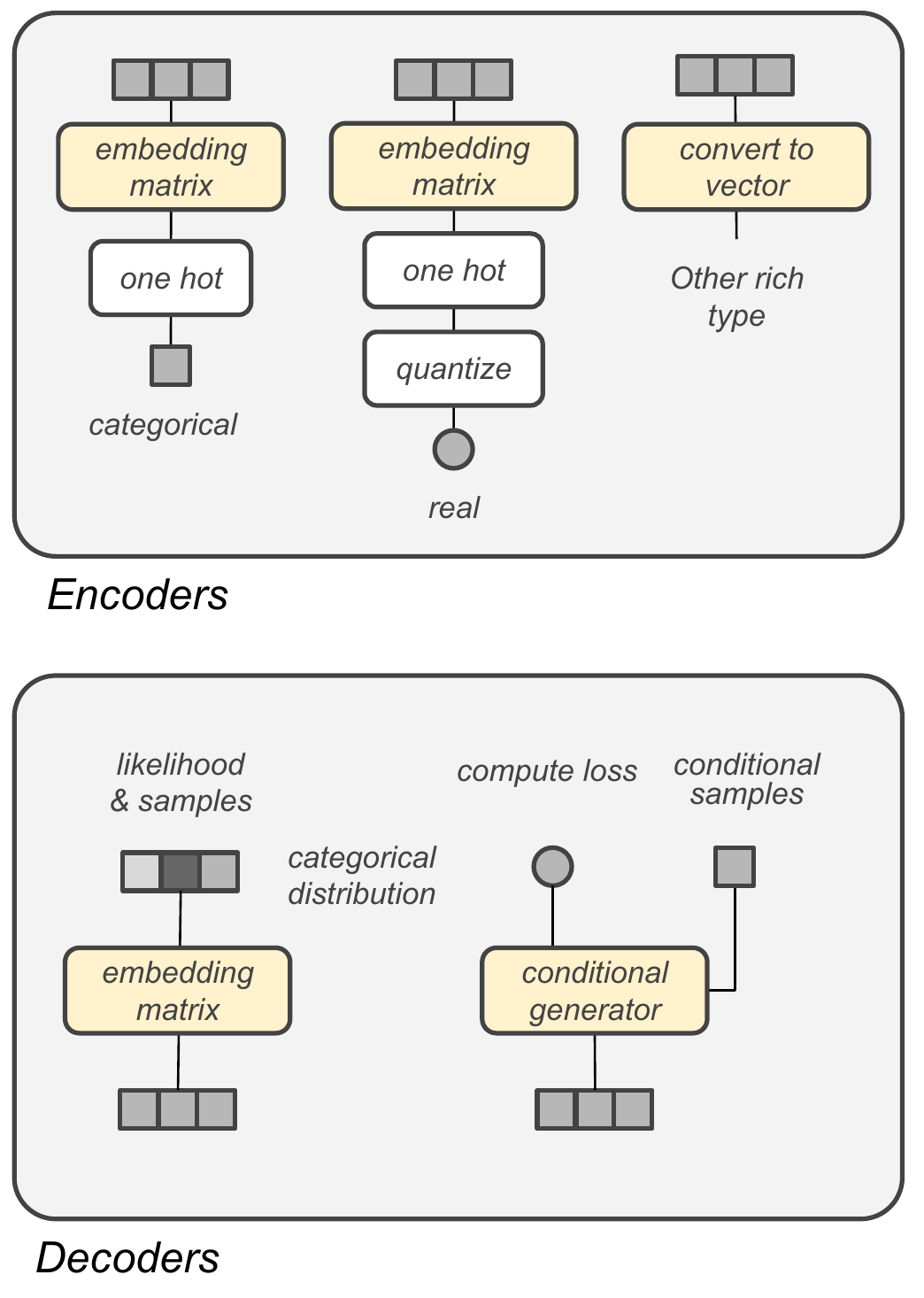}
        \caption{Encoder and decoder detail} \label{fig:1c}
    \end{subfigure}
    \caption{Overview of the model architecture.} \label{fig:1}
\end{figure}

% Tabular - flexible - missing values.
% Se caler plus sur: [Xu et al. 2019 - Modeling Tabular Data using Conditional GAN](https://arxiv.org/pdf/1907.00503.pdf)
% Or https://arxiv.org/pdf/2012.06678.pdf

% TODO
% [x] Intro NG
% [x] Related work
%  [x] Tabular + un peu multimodal JL
%  [x] Missing values <- multimodal NG
% [x] Refactor model
% [x] Change the notion of 'positional embedding' to 'column embedding' (or a better name if you have an idea)
% [x] Metadata (not so short to do)
% [x] Nice biblio (unshort it)
% [x] Codec requirements -> exemple
% [x] Deployment -> conclusion
% [x] Overfitted models ?? The likelihood of the generated data is not optimized by the model ??? JL Fig
% [x] Model desc not so clear JL autoporteur + shuffling -> fig + JL
% [x] Pas de caption acc, f1, Lik etc. JL
% [x] Loss JL
% [x] Refs
% [x] Conclusion

% Deployed in production:
% http://opencesp.vjf.inserm.fr/sarus/
% https://sarus.tech/

% https://www.kdd.org/kdd2021/calls/view/call-for-applied-data-science-track-papers

\section{Introduction}

There have recently been impressive advances in \emph{deep generative modeling}
techniques, which have made new industrial applications possible.

These include new ways of collaborating on sensitive data with
strong privacy guarantees such as the release of synthetic microdata
by the \emph{US Census Bureau} \cite{benedetto2018creation}.
The idea behind this application of synthetic data, is that training a generative model
with \emph{differential privacy} \cite{abadi2016deep, jordon2018pate}, and then
sharing the data generated from the model, enables the sharing of statistical properties
of privacy-sensitive data, without sharing the data itself.

There is a long line of work about \emph{differential privacy} \cite{dwork2006calibrating, dwork2014algorithmic},
which establishes it as a solid notion of privacy. This paper will not
develop this aspect, but rather focus on generative modelling in this particular context.

A very common use-case for generating synthetic data, is the need to share
privacy sensitive \emph{tabular data} mixing together various column types
such as numerical, categorical, or \emph{richer types} like sequences or images.

Furthermore, in training a model with \emph{differentially private} techniques such
as \emph{DP-SGD} \cite{abadi2016deep}, one generally wants to interact as little as
possible with the private data to minimize privacy loss, and therefore one wants to pre-train
the model with public datasets. In practice, these public datasets often have partially
overlapping sets of features.
To be pretrained on many of these heterogenous datasets, a synthetic data model needs
to handle properly missing-data in a way similar to semi-supervised learning \cite{kingma2014semi}.

This paper proposes a novel deep generative model architecture (see figure~\ref{fig:1}), the
\emph{\modelfull{}} (\model), meeting the requirements specific to practical
privacy preserving synthetic data generation, namely:
\begin{itemize}
    \item to accomodate a variety of column types, continuous, discrete and potentially richer types by \emph{composing} them
    \item to handle missing data properly
    \item to show good performances in modelling the joint distribution of data
\end{itemize}

The proposed architecture is autoregressive in the sense that each feature
of the data is generated sequentially conditional on the features already
generated. The order in which features are generated does not mater.

During training, the features are shuffled and each of them is encoded in a separate
\emph{input representation} vector to effectively bypass the bottleneck of encoding a variable length input
into a fixed size vector \cite{cho2015describing}.

The input representations are then combined with \emph{column embeddings} and
passed through a causal transformer \cite{vaswani2017attention} that builds
a set of higher level representations in a potentially generalized many-to-many
cross-feature inference.

An \emph{output representation} of a feature is built from the input
representations of one or several other features and the \emph{embedding}
of the generated feature. The output representation for a feature sums up
the shared knowledge of one or more other features.
It is fed to a conditional generative model specialized
for this feature (e.g. CNNs for images, RNNs for sequences, etc).
These conditional generative models can be off-the-shelf pre-trained modules plugged
into our framework. An overview of the framework is pictured on
figure~\ref{fig:1}.

\textbf{This work provides the following contributions}:
\begin{itemize}
    \item \emph{We describe a novel architecture} built upon the Transformer, using \emph{column embeddings}
    to encode the feature position and \emph{shuffling features during training} to learn all
    possible combinations of missing values.
    \item Following \cite{xu2019modeling}, we evaluate our architecture on 13 datasets (6 standard datasets and 7 simulated),
    compare it to 14 other generative models and demonstrate \emph{it beats the state of the art in tabular data
    generation}
    \item We specify a set of requirements any conditional generative model needs to implement
    to be used as a component of the \model{}, \emph{enabling the use of richer datatypes} than numerical or categorical.
\end{itemize}

\section{Related Work}

\subsection{Tabular data generation}
Tabular data are very widespread in the industry. Yet relatively few data
generation research focuses on tabular data. Traditional models for tabular data
generations use decision trees, bayesian networks or copulas to sample from a
data distribution.

Recent approaches have used GANs to generate tabular data. Some of them have
focused their efforts on generating Electronic Health Records (EHR).
MedGAN~\cite{choi2017generating} pretrains an autoencoder and then uses the
decoder as the final part of the generator, tableGAN~\cite{park2018data}
introduces convolutions in the generator, PATE-GAN~\cite{jordon2018pate} uses
the PATE framework to generate differentially private synthetic data while CTGAN
and TVAE~\cite{xu2019modeling} introduce a node specific normalization and
conditional sampling to tackle mode collapse. The last paper also introduced a
useful benchmark framework called SDGym
\footnote{https://github.com/sdv-dev/SDGym} for comparing tabular data
generative models.

\subsection{Composable generative models} % Finish that and explain the challenge of missing values

Beyond the focus on tabular data, we designed our architecture so that it could be
trained on any subset of features without having to train an exponential number of models
(one for each possible subset).

Some early attempts in this direction were based on Restricted Boltzmann Machines (RBM)
\cite{srivastava2014multimodal}.

An important line of work is based on the variational auto-encoder (VAE).
Composable VAEs aim to provide a flexible mechanism to
compose generative models and adapt to arbitrary missing values patterns without having
to learn an exponential number of mapping networks.
MVAE~\cite{wu2018multimodal}, MMVAE~\cite{shi2019variational},
mmJSD~\cite{sutter2020multimodal} and MHVAE~\cite{vasco2020mhvae} propose to
have one specialized encoder per modality (called expert) and a rule for
combining the experts' outputs together to infer the latent variable.

Our proposed approach leverages the \emph{Transformer} architecture to combine
several conditional generator sub-models.

\subsection{Transformers}
The Transformer \cite{vaswani2017attention} has been shown to excel in several
modalities such as natural language \cite{devlin2018bert, radford2019language, brown2020language},
image \cite{parmar2018image,dosovitskiy2020image} or music
\cite{huang2018music}. As each self-attention layer has a global receptive
field, the network can give more importance to the input regions most useful for
predicting a given point. Thus the architecture may be more flexible at
generating diverse data types than networks with fixed connectivity patterns
\cite{child2019generating}.

At the core of the Transformer is the \emph{attention} operation which can be seen as a
list of queries on a set of key and value pairs. We say that the attention is
causal if the $k$th representation vector depends only on the $k-1$ first
values. Considering a sequence of length $l$, a (key, query) dimension $d_k$ and
a value dimension $d_v$, we can define the \emph{attention} and
\emph{causal attention} operations as follows:

\begin{equation} \label{eq2}
    \begin{split}
        \text{attention}(Q,K,V) & = \text{softmax}\left(\frac{Q K^T}{\sqrt{d_k}}\right) V^T \\
        \text{causal\_attention}(Q,K,V) & = \text{softmax}\left(\frac{Q K^T}{\sqrt{d_k}}-M\right) V^T
    \end{split}
\end{equation}

Where $Q$ and $K$ have shape $(l, d_k)$, $V$ has shape $(l,d_v)$ and
where the mask $M$ is a lower triangular matrix with shape $(l, l)$ where
$M_{ij}=1_{j\leq i} \infty$, which enforces the causality by setting
contributions of input $j$ to output $i$ to $0$ if $j>i$.

Following \cite{vaswani2017attention, radford2019language, brown2020language},
we will name \emph{self-attention} (or \emph{causal-self-attention}) the use of a multi-head \emph{attention}
(or \emph{causal-attention}) operation with linear transformations of the same vector passed as $Q$, $K$ and $V$ arguments.

And we will name \emph{cross-attention} (or \emph{causal-cross-attention}) the use of a multi-head \emph{attention}
(or \emph{causal-attention}) operation with linear transformations of the same vector passed as $K$ and $V$ arguments and
linear transformations of another vector as $Q$.

\section{The \modelfull{}}

\subsection{Model definition}

Let us consider a tabular dataset consisting of categorical and real values.
Cells can also contain missing values. In all this section, we will encode real
values as categorical variables by quantizing it using its quantiles. We could
also use \emph{mode aware encoding} techniques as described in
\cite{xu2019modeling}.

Formally, we define our dataset $\mathcal{D}$ as a set of $n$ \emph{features}
(or columns) $\mathcal{F}_k$ where the $k$th feature is $d_k$ dimensional i.e.
$\mathcal{F}_k \in \mathbb{R}^{d_k}$. The i$th$ training example consists of
values $\mathcal{F}^i = \{f_1^i, ..., f_n^i\}$. Assuming training examples are
i.i.d. realization of a random variable with distribution $\mathcal{F}^i\sim
\mathcal{P_D}$, our objective is to train a model that can sample from an
estimator $\mathcal{\hat{P}_D}$ maximizing the likelihood of the data.

We introduce an architecture built around a Transformer stack. We use the same
transformer stack architecture as GPT-2 and GPT-3
\cite{radford2019language, brown2020language}. The Transformer builds a sequence
of representations from the input values.

For a training example $i$ and feature $k$, our model is formally defined by

\begin{equation} \label{eq1}
    \begin{split}
        \mathcal{R}^i & = \text{causal\_transformer}(E(\mathcal{F}^i) + \mathcal{X}) \\
        y_k^i & = \text{cross\_attention}(Q=x_k^i, (K,V)=\mathcal{R}_{<k}^i)
    \end{split}
\end{equation}

Since each feature $f_k^i$ has a priori a different dimension (i.e. number of
categories), we first need to project the values to a common latent subspace so
that they can then be fed into the transformer (in the same way as words are
represented as dense vector embeddings). For this purpose, each feature
$\mathcal{F}_k$ has an \emph{encoder} $E_k$ that converts its data space to the
latent space $E_k\left(f_k^i\right) \in \mathbb{R}^h$.

Furthermore, as there is no natural ordering of features, for each feature
$\mathcal{F}_k$ we learn a \emph{column embedding} $x_k \in \mathbb{R}^h$ that
is integrated to the input by adding it to the latent value. The column
embedding can be thought of as an origin point of the feature in the latent
space $\{E_1(f_1^i)+ x_1, ..., E_n(f_n^i) + x_n\} = E(\mathcal{F}^i) +
\mathcal{X}$. The column embedding is learned and identifies the columns.

The \emph{causal-Transformer} is composed of a causal-self-attention operation
and a position-wise feedforward network both preceded by a normalization layer
and followed by a residual connection. The feedforward network has the following
structure where, if $X \in \mathbb{R}^d$, $W_1$ has shape $(4d,d)$ and $W_2$ has
shape $(d,4d)$:

\begin{equation} \label{eq5}
    \begin{split}
        \text{feedforward}(X)=W_2.\text{gelu}(W_1.X)
    \end{split}
\end{equation}

\subsection{Conditional distribution}

At the output of the Transformer, the representation vector $y_k$ summarizes the
distribution of the $k^{th}$ feature conditionally on the previous features:

\begin{equation} \label{eq6}
    \mathcal{P}(f_k|f_1,...,f_{k-1}) = \mathcal{P}(f_k|y_k)
\end{equation}

For each feature, we add a decoder $D_k$ that could be any off-the-shelf
conditional generation model and trained accordingly. The decoder $D_k$ allows
us to compare the conditional distribution to the expected data distribution.
The provided error signal for each feature is backpropagated to train the whole
model.

In this paper, we only consider the case of the conditional categorical
generative model. For this purpose, we reuse the embedding matrix $E_k$ used to
transform the one-hot encoded vectors to dense latent vactor. We multiply the
conditional vector $y_k \in \mathbb{R}^h$ with the transpose of the embedding matrix $E_k \in \mathbb{R}^{(h, d_k)}$ to
produce an $n_k$ dimentional vector representing the logits of the categorical
distribution $\mathcal{P}(f_k|y_k)$, where $d_k$ is the number of caterogies (or
quantiles) of the $k^{th}$ feature.

\begin{equation} \label{eq7}
    \mathcal{P}(f_k|y_k) = \text{softmax}(E_k^Ty_k)
\end{equation}

The model then minimize the loss $\mathcal{L}$ which is the categorical
cross-entropy accross all columns as written in equation \ref{eq8} where $f_k$
is a one-hot encoded vector.

\begin{equation} \label{eq8}
    \mathcal{L} = - \sum_k f_k \log(\text{softmax}(E_k^T y_k))
\end{equation}

\subsection{Training \model{}s}

\paragraph{Feature order} One of the advantages of this model compared to a
traditional architecture \cite{uria2016neural} is the fact that \emph{the
autoregressive order does not need to be fixed}. For each training example, a
random permutation of the input features is chosen and the probability
distribution of the $k^{th}$ feature conditionally on a random subset of
the remaining features is computed. The training algorithm is summarized in
algorithm~\ref{algo:training}.

\begin{figure}[!htb]
    \centering
    \includegraphics[width=\linewidth]{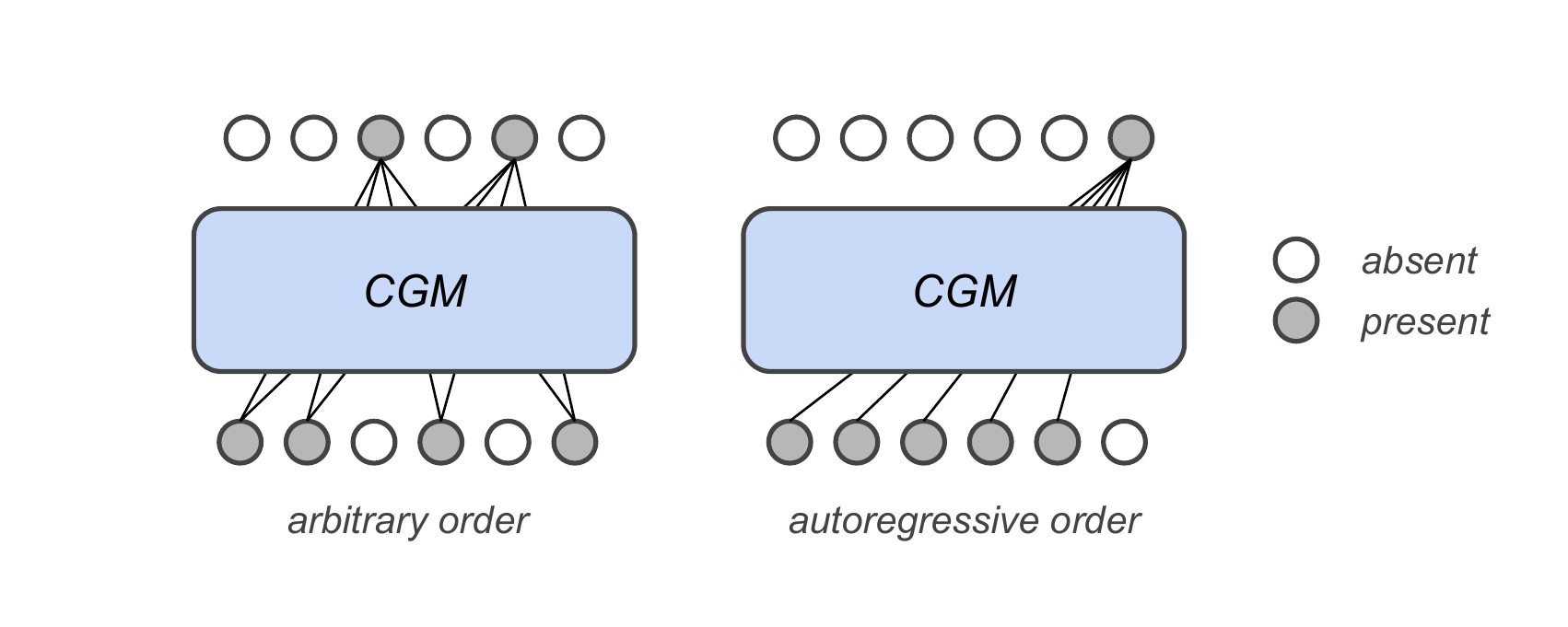}
    \caption{\model{} can use any set of columns to make predictions.}
    \label{fig:composability}
\end{figure}

\paragraph{Weakly-supervised setting} Another advantage of our model is its
ability to be trained in a weakly-supervised setting, that is, when some columns
have missing values. This is possible since the columns are referred to by
explicitly learned column embedding as shown in figure~\ref{fig:composability}.

\begin{algorithm}[h]
\SetAlgoLined Initialize $\theta_E, \theta_T, \theta_D$\;
\While{not converged}{
    $\mathcal{H} \leftarrow E(\mathcal{F}) + \mathcal{X}$\;
    $\mathcal{H} \leftarrow \text{permute}(\mathcal{H})$\;
    $\mathcal{R} \leftarrow \text{causal\_transformer}(\mathcal{H})$\;
    $y_k \leftarrow \text{cross\_attention}(Q=x_k, (K, V)=\mathcal{R}_{<k}), \forall k$\;
    $\mathcal{L} \leftarrow - \sum_k f_k \log(\text{softmax}(E_k^T y_k))$\;
    $\theta_E, \theta_T, \theta_D \leftarrow (\theta_E, \theta_T, \theta_D) -
    \lambda \nabla_{\theta_D, \theta_T, \theta_D} \mathcal{L}$\;}
    \caption{Training algorithm}
    \label{algo:training}
\end{algorithm}

\subsection{Generation with \model{}s}

Once trained, the model can be used for generation. The first representation
vector is generated using an empty set of features. The
first features is then sampled from the representation.
The other features are generated conditionally on the
ones already generated:
\begin{equation} \label{generation}
    \begin{split}
        \mathcal{R} & = \text{causal\_transformer}(E(\mathcal{F}) + \mathcal{X}) \\
        y_k & = \text{cross\_attention}(Q=x_k, (K,V)=\mathcal{R}_{<k}) \\
        f_k & \sim \mathcal{P}(f_k|y_k)
    \end{split}
\end{equation}

% Show any model fulfilling the requirements can be used as a type codec

\section{Composing a model with richer types}

Because it is trained with \emph{feature shuffling}, it is
possible to remove or add features to the \model{} at any stage of the training.
Those features can be numerical or categorical as described above, but they could
be of a richer type provided the type implements the interface specified below:
\begin{description}
    \item[Encoder] the type of feature $\mathcal{F}_k$ should be equiped with an
    encoder, mapping a value $f_k^i$ from the data to an \emph{input representation}:
    $E_k:  \to \mathbb{R}^h$

    The encoder can be a neural network, parametrized by weights and trained along with those of the
    transformer.

    \item[Decoder] the type of feature $\mathcal{F}_k$ should be able equiped
    with a decoder, sampling a value $f_k \in \mathbb{R}^{d_k}$ from an \emph{output representation}
    $y_k in \mathbb{R}^h$: $f_k \sim \mathcal{P}(f_k|y_k)$

    The decoder can be a neural network, parametrized by weights and trained along with those of the
    transformer.

    \item[Loss] the type of feature $\mathcal{F}_k$ should provide the loss to minimize during training
    of the weights of the transformer along with \emph{encoder weights} and \emph{decoder weights}.

    The loss can also be parametrized as a neural network and trained against an adversarial loss, thus
    permiting the integration of \emph{Generative Adversarial Networks} \cite{goodfellow2014generative}
    as the decoder part of a rich type (like images or sound).
\end{description}

When considering a \emph{categorical} feature, the \emph{encoder} is simply parametrized by an
embedding matrix giving the \emph{input representation} vector corresponding to the $i^{th}$ modality
by a simple look-up operation.
The \emph{decoder} is also parametrized by an embedding matrix converting an \emph{output representation}
vector to a vectors of logits fed into a softmax. Then a value is sampled from the probabilities derived
from the softmax.
The \emph{loss} is simply the categorical cross-entropy.

\emph{Numerical} feature are converted to categorical by quantization. The \emph{encoder} and \emph{decoder}
have the same form as those of a categorical feature, simply composed with a quantization, de-quantization
step.

If one wants to produce synthetic data with columns of images, one can simply implement the specifications
above. The \emph{encoder} can be a deep convolutional network, the \emph{decoder} the generator network
of a GAN and the \emph{loss} the dicriminator network of the GAN. In that case two adversarial
objectives are optimized during training, the minimization of the common loss of all the encoders, decoders
and transformer networks, and the minimization of the adversarial loss of the discriminators of image features.

Beyond images, one can integrate sequences usng RNN or Transformers, NLP and more. The setting with images
categorical and numerical features has been tested and works fine. And further types are being integrated
in the model.

\section{Application to tabular data}

In this section we present the experiments we conducted to test the performance
and applications of our model on tabular data.

% \begin{figure}[!htb]
%     \centering
%     \includegraphics[width=\linewidth]{sdgym_overview.pdf}
%     \caption{Overview of the SDGym scoring process \cite{xu2019modeling}.}
%     \label{fig:sdgym}
% \end{figure}

\begin{figure*}
    \centering
    \begin{subfigure}{0.35\textwidth}
        \includegraphics[width=\linewidth]{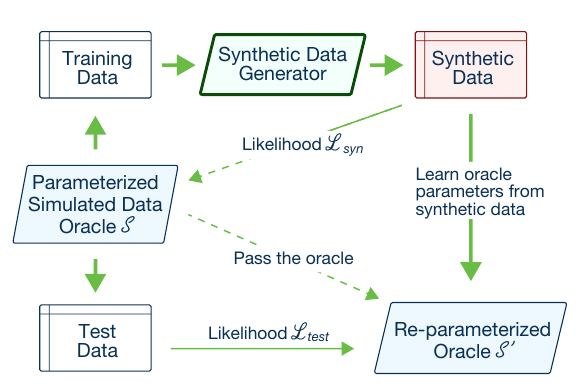}
        \caption{SDGym process for synthetic datasets.} \label{fig:sdgym_syn}
    \end{subfigure}
    % \hspace*{\fill}
    \begin{subfigure}{0.4\textwidth}
        \includegraphics[width=\linewidth]{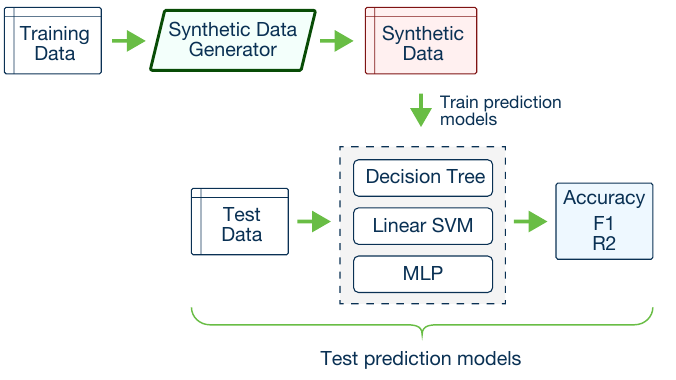}
        \caption{SDGym process for real datasets.} \label{fig:sdgym_real}
    \end{subfigure}
    \caption{Overview of the SDGym scoring process \cite{xu2019modeling}.}
    \label{fig:sdgym_overview}
\end{figure*}

\begin{figure*}[!htb]
    \centering
    \begin{tabular}{lllllllllll}
\toprule
{} & \multicolumn{2}{l}{adult} & \multicolumn{2}{l}{census} & \multicolumn{2}{l}{covtype} & \multicolumn{2}{l}{credit} & \multicolumn{2}{l}{intrusion} \\
{} &       acc &             f1 &                    acc &             f1 &       acc &       f1 &       acc &                          f1 &       acc &       f1 \\
\midrule
\texttt{Uniform}                &           0.49 &           0.24 &                        0.54 &           0.11 &           0.14 &           0.09 &           0.57 &                        0.01 &           0.12 &           0.07 \\
\texttt{Independent}            &           0.64 &           0.15 &                        0.66 &           0.06 &           0.38 &           0.11 &           0.88 &                        0.00 &           0.72 &           0.20 \\
\texttt{GCCF}                   &           0.77 &           0.26 &                        0.79 &           0.05 &           0.42 &           0.14 &           0.98 &                        0.01 &           0.84 &           0.26 \\
\texttt{GCOH}                   &           0.78 &           0.20 &  \color{red}{\textbf{0.93}} &           0.13 &           0.51 &           0.18 &  \textbf{1.00} &                        0.00 &           0.90 &           0.33 \\
\texttt{CopulaGAN}              &           0.79 &           0.61 &                        0.89 &           0.39 &           0.57 &           0.33 &  \textbf{1.00} &  \color{red}{\textbf{0.72}} &           0.98 &           0.52 \\
\texttt{CLBN}                   &           0.77 &           0.31 &                        0.90 &           0.29 &           0.56 &           0.33 &  \textbf{1.00} &                        0.44 &           0.94 &           0.39 \\
\texttt{PrivBN}                 &           0.80 &           0.43 &                        0.90 &           0.24 &           0.47 &           0.21 &           0.96 &                        0.01 &           0.95 &           0.38 \\
\texttt{Medgan}                 &           0.62 &           0.24 &                        0.64 &           0.14 &           0.44 &           0.09 &           0.94 &                        0.02 &           0.87 &           0.27 \\
\texttt{VEEGAN}                 &           0.72 &           0.16 &                        0.76 &           0.05 &           0.22 &           0.10 &           0.88 &                        0.17 &           0.51 &           0.18 \\
\texttt{TVAE}                   &           0.80 &           0.62 &  \color{red}{\textbf{0.93}} &           0.38 &           0.65 &           0.46 &  \textbf{1.00} &                        0.00 &           0.97 &           0.43 \\
\texttt{CTGAN}                  &           0.80 &           0.60 &                        0.90 &           0.38 &           0.58 &           0.33 &           0.99 &                        0.52 &           0.98 &           0.51 \\
\texttt{\textbf{\model (ours)}} &  \textbf{0.83} &  \textbf{0.65} &               \textbf{0.91} &  \textbf{0.41} &  \textbf{0.70} &  \textbf{0.49} &  \textbf{1.00} &               \textbf{0.53} &  \textbf{0.99} &  \textbf{0.59} \\
\texttt{Identity}               &           0.82 &           0.66 &                        0.91 &           0.46 &           0.76 &           0.65 &           0.99 &                        0.55 &           1.00 &           0.86 \\
\bottomrule
\end{tabular}

    \caption{Real datasets leaderboard. Each synthesizer is trained on a training split $\mathcal{T}_{train}$ of the real data. Synthetic data $\mathcal{T}_{syn}$ are generated and a classifier is trained on it. The classifier's accuracy and f1-score are then measured on a test split $\mathcal{T}_{test}$  of the real data.}
    \label{fig:real}
\end{figure*}

\begin{figure*}[!htb]
    \centering
    \begin{tabular}{lllllll}
\toprule
{} & \multicolumn{2}{l}{grid} & \multicolumn{2}{l}{gridr} & \multicolumn{2}{l}{ring} \\
{} &               $\mathcal{L}_{syn}$ & $\mathcal{L}_{test}$ &  $\mathcal{L}_{syn}$ & $\mathcal{L}_{test}$ &               $\mathcal{L}_{syn}$ & $\mathcal{L}_{test}$ \\
\midrule
\texttt{Uniform}                &                        -7.33 &           -4.54 &           -7.22 &           -4.57 &                        -5.34 &           -2.50 \\
\texttt{Independent}            &               \textbf{-3.47} &  \textbf{-3.49} &           -5.12 &           -4.03 &                        -2.47 &           -1.96 \\
\texttt{GCC}                    &                        -7.24 &           -4.51 &           -7.16 &           -4.54 &                        -3.20 &           -2.15 \\
\texttt{GCCF}                   &                        -7.34 &           -4.57 &           -7.14 &           -4.55 &                        -3.18 &           -2.15 \\
\texttt{GCOH}                   &                        -7.27 &           -4.51 &           -7.19 &           -4.55 &                        -3.21 &           -2.15 \\
\texttt{CopulaGAN}              &                        -8.19 &           -5.14 &           -8.16 &           -5.01 &                        -6.21 &           -2.80 \\
\texttt{CLBN}                   &                        -3.88 &           -9.20 &           -4.01 &           -7.43 &               \textbf{-1.77} &          -47.16 \\
\texttt{Medgan}                 &                        -5.83 &          -90.34 &           -7.37 &         -141.41 &                        -2.78 &         -149.77 \\
\texttt{VEEGAN}                 &                        -8.65 &         -423.57 &          -11.46 &           -8.91 &                       -16.83 &           -6.35 \\
\texttt{Tablegan}               &                        -6.99 &           -5.33 &           -7.00 &           -4.83 &                        -4.74 &           -2.53 \\
\texttt{TVAE}                   &  \color{red}{\textbf{-3.27}} &           -5.66 &  \textbf{-3.87} &  \textbf{-3.71} &  \color{red}{\textbf{-1.58}} &           -1.94 \\
\texttt{CTGAN}                  &                        -8.92 &           -5.09 &           -8.32 &           -5.03 &                        -7.13 &           -2.70 \\
\texttt{\textbf{\model (ours)}} &                        -3.50 &           -3.54 &           -3.89 &  \textbf{-3.71} &                        -1.82 &  \textbf{-1.74} \\
\texttt{Identity}               &                        -3.47 &           -3.49 &           -3.59 &           -3.64 &                        -1.71 &           -1.70 \\
\bottomrule
\end{tabular}

    \caption{Synthetic datasets leaderboard. A parametric model $\mathcal{M}$ is used to generate training data $\mathcal{T}_{train}$ and test data $\mathcal{T}_{test}$. Each synthesizer is trained and generates synthetic data $\mathcal{T}_{syn}$. The log-likelihood of the synthetic data wrt. $\mathcal{M}$ is computed $\mathcal{L}_{syn} = log(P(\mathcal{T}_{syn}|\mathcal{M}))$. The parametric model is also refitted using $\mathcal{T}_{syn}$ and yield a new model $\mathcal{M}'$. The log-likelihood of the test data wrt. $\mathcal{M}'$ is computed $\mathcal{L}_{test} = log(P(\mathcal{T}_{test}|\mathcal{M}'))$.}
    \label{fig:synthetic}
\end{figure*}

\begin{figure*}[!htb]
    \centering
    \begin{tabular}{lllllllll}
\toprule
{} & \multicolumn{2}{l}{alarm} & \multicolumn{2}{l}{asia} & \multicolumn{2}{l}{child} & \multicolumn{2}{l}{insurance} \\
{} &               $\mathcal{L}_{syn}$ &  $\mathcal{L}_{test}$ &               $\mathcal{L}_{syn}$ & $\mathcal{L}_{test}$ &                $\mathcal{L}_{syn}$ &  $\mathcal{L}_{test}$ &   $\mathcal{L}_{syn}$ &  $\mathcal{L}_{test}$ \\
\midrule
\texttt{Uniform}                &                       -18.42 &           -18.42 &                       -14.28 &           -5.52 &                        -18.41 &           -18.32 &           -18.42 &           -18.42 \\
\texttt{Independent}            &                       -18.24 &           -15.82 &                        -4.97 &           -2.99 &                        -17.96 &           -16.03 &           -18.38 &           -17.57 \\
\texttt{GCC}                    &                       -12.91 &           -15.57 &               \textbf{-2.25} &           -3.61 &                        -16.40 &           -15.55 &           -17.84 &           -16.58 \\
\texttt{GCCF}                   &                       -14.52 &           -14.57 &                        -2.83 &           -3.11 &                        -16.90 &           -15.41 &           -18.03 &           -16.52 \\
\texttt{GCOH}                   &                       -15.48 &           -15.67 &                        -2.31 &           -3.23 &                        -14.48 &           -15.31 &           -17.84 &           -17.91 \\
\texttt{CopulaGAN}              &                       -15.69 &           -13.05 &                        -3.96 &           -2.40 &                        -14.25 &           -12.92 &           -16.96 &           -14.96 \\
\texttt{CLBN}                   &                       -12.46 &           -11.19 &                        -2.40 &           -2.27 &                        -12.63 &           -12.31 &           -15.17 &           -13.92 \\
\texttt{PrivBN}                 &                       -12.15 &           -11.14 &                        -2.29 &  \textbf{-2.24} &               \textbf{-12.36} &           -12.19 &           -14.70 &           -13.64 \\
\texttt{Medgan}                 &  \color{red}{\textbf{-7.84}} &           -13.26 &  \color{red}{\textbf{-1.57}} &           -5.97 &  \color{red}{\textbf{-11.11}} &           -12.99 &           -13.88 &           -15.08 \\
\texttt{VEEGAN}                 &                       -18.39 &           -18.21 &                       -11.49 &           -5.95 &                        -17.31 &           -17.68 &           -18.33 &           -18.11 \\
\texttt{Tablegan}               &                       -12.69 &           -11.54 &                        -3.40 &           -2.72 &                        -15.02 &           -13.39 &           -16.18 &           -14.32 \\
\texttt{TVAE}                   &                       -11.44 &           -10.76 &                        -2.29 &           -2.27 &                        -12.46 &           -12.30 &           -14.30 &           -14.24 \\
\texttt{CTGAN}                  &                       -15.22 &           -12.93 &                        -2.69 &           -2.31 &                        -13.81 &           -12.81 &           -16.60 &           -14.84 \\
\texttt{\textbf{\model (ours)}} &              \textbf{-10.63} &  \textbf{-10.48} &                        -2.33 &           -2.25 &  \color{red}{\textbf{-11.83}} &  \textbf{-12.13} &  \textbf{-13.21} &  \textbf{-13.11} \\
\texttt{Identity}               &                       -10.23 &           -10.30 &                        -2.24 &           -2.24 &                        -12.03 &           -12.04 &           -12.85 &           -12.96 \\
\bottomrule
\end{tabular}

    \caption{Bayesian datasets leaderboard. A parametric model $\mathcal{M}$ is used to generate training data $\mathcal{T}_{train}$ and test data $\mathcal{T}_{test}$. Each synthesizer is trained and generates synthetic data $\mathcal{T}_{syn}$. The log-likelihood of the synthetic data wrt. $\mathcal{M}$ is computed $\mathcal{L}_{syn} = log(P(\mathcal{T}_{syn}|\mathcal{M}))$. The parametric model is also refitted using $\mathcal{T}_{syn}$ and yield a new model $\mathcal{M}'$. The log-likelihood of the test data wrt. $\mathcal{M}'$ is computed $\mathcal{L}_{test} = log(P(\mathcal{T}_{test}|\mathcal{M}'))$.}
    \label{fig:bayesian}
\end{figure*}

We trained the model on tabular data and benchmarked against the SDGym
leaderboard. SDGym evaluates the performance of synthetic data generators on
three dataset families: simulated data using gaussian mixtures, simulated data
using bayesian networks and real world datasets \cite{xu2019modeling}.

We evaluated our model using the publicly available SDGym benchmark
\cite{xu2019modeling}. The model's performance at generating synthetic data is
measured against several other models on different types of datasets.

The test procedure is different for datasets generated with a parametric model
$\mathcal{M}$ (synthetic and bayesian) and real datasets as seen on
figure~\ref{fig:sdgym_overview}. For each dataset, the data is split into a
training set $\mathcal{T}_{train}$ and a test set $\mathcal{T}_{test}$. The
generative model is trained on $\mathcal{T}_{train}$ and a synthetic data set
$\mathcal{T}_{syn}$ is generated.

\paragraph{Synthetic and bayesian datasets} For generated datasets (synthetic
and bayesian datasets), the probability distribution of the data is known. We
can thus evaluate the log-likelihood of the synthetized data with respect to the
parametric model $\mathcal{M}$ that is $\mathcal{L}_{syn} =
log(P(\mathcal{T}_{syn}|\mathcal{M}))$. However, such a metric favors mode
collapsed models. To measure if mode collapse has occured, the parametric model
$\mathcal{M}$ that was used to generate $\mathcal{T}_{train}$ is refitted using
$\mathcal{T}_{syn}$ instead. This yields a new parametric model $\mathcal{M}'$.
The log-likelihood of the test set $\mathcal{T}_{test}$ with respect to
$\mathcal{M}'$ is then computed $\mathcal{L}_{test} =
log(P(\mathcal{T}_{test}|\mathcal{M}'))$.

\paragraph{Real datasets} For real datasets, the machine learning efficacy is
used to measure the quality of the synthetic data. A classification model is
trained using $\mathcal{T}_{syn}$ and the accuracy $acc$ and $f1$ scores are
measured on the test set $\mathcal{T}_{test}$.

For all datasets, we used a hidden dimension of 64 and a transformer composed of
2 blocks with 8 heads. We trained the generative model for 15 epochs with a
batch size of $128$ and the Adam optimizer with $\beta_1=0.5$, $\beta_2=0.99$, a
learning rate of $0.001$. All computations were done on a machine with 8 CPUs
and 2 NVIDIA V100 GPUs. Benchmarking on all SDGym datasets took 4 hours.

\subsection{Results}
We compare our results with the public leaderboard provided on the SDGym web
page. The scores of several models have been pre-computed for comparison. We are
compared against the following models: \texttt{GCC} (Gaussian Copula
Categorical), \texttt{GCCF} (Gaussian Copula Categorical Fuzzy), \texttt{GCOH}
(Gaussian Copula One Hot), \texttt{CopulaGAN}, \texttt{CLBN}, \texttt{PrivBN},
\texttt{Medgan}, \texttt{Tablegan}, \texttt{CTGAN} and \texttt{TVAE}.

Trivial synthesizers are also benchmarked: the \texttt{Uniform} synthesizer
generates data uniformly and serves as a lower bound likelihood estimator, the
\texttt{Independent} synthesizer makes an independence assumption and the
\texttt{Identity} synthesizer simply returns the training data and serves as a
likelihood upper bound estimator. The actual scores are presented on figures
\ref{fig:real}, \ref{fig:synthetic} and \ref{fig:bayesian}.

For model comparison, we will focus on the $\mathcal{L}_{test}$ and f1-scores
since the other metrics (accuracy and $\mathcal{L}_{syn}$) do not reflect the
synthetic data quality as good. To easily visualize the results, we put in
\textbf{bold} the best result in each column.

As we can see, \textbf{our model outperforms the other models on the real
datasets}. On the credit dataset, we can see that the \texttt{CopulaGAN} model
outperforms us but that is also outperforms the \texttt{Identity} synthesizer by
a large margin. Such a high performance, better than the perfect
\texttt{Identity} synthesizer, is odd and could be attributed to the fact that
the synthesized data are better suited to train the subsequent classifier
whereas the real data may contain outliers that perturb the classifier's
training process.

On synthetic datasets, our model is the leader in the outperformed
\texttt{gridr} and \texttt{ring} datasets. It is slightly outperformed by the
trivial \texttt{Independent} synthesizer on the \texttt{grid} dataset but is the
second best model by a large margin.

On bayesian datasets, our model is the leader on the \texttt{alarm},
\texttt{asia} and \texttt{insurance} datasets. Our model is very close to the
\texttt{PrivBN} synthesizer on the \texttt{asia} dataset though slightly
outperformed.

\section{The \model{} in production}

The \model{} is used in production at \href{https://sarus.tech}{Sarus Technologies}
for the generation of synthetic data.

\href{https://sarus.tech}{Sarus} is a technology company developing solutions
to analyze privacy-sensitive data with formal privacy guarantees.

In production, the \model{} is trained with \emph{DP-SGD} \cite{abadi2016deep} in order to provide
differential privacy guarantees to the data generator.

The solution has been applied in medical data sharing with the
\href{https://cesp.inserm.fr/}{French National Institute in Medical Research (INSERM)}
and it is being experimented in various institutions in finance, energy, and transportation.

\section{Conclusion}

Though we focus on categorical and numerical data in this paper, the \model{} can accomodate a wide range of types.

Besides, as demonstrated, the \model{} is able to be pre-trained on many public datasets before
it is trained on private data.

Both those improvements and the quality of the data generated,
open new possibilities in privacy preserving data analysis, enabling
new applications in health care, personal finance or public policy planning.

\bibliographystyle{alpha}
\bibliography{multimodal_data}

\newcommand{\etalchar}[1]{$^{#1}$}
\begin{thebibliography}{GPAM{\etalchar{+}}14}

\bibitem[ACG{\etalchar{+}}16]{abadi2016deep}
Martin Abadi, Andy Chu, Ian Goodfellow, H~Brendan McMahan, Ilya Mironov, Kunal
  Talwar, and Li~Zhang.
\newblock Deep learning with differential privacy.
\newblock In {\em Proceedings of the 2016 ACM SIGSAC conference on computer and
  communications security}, pages 308--318, 2016.

\bibitem[BMR{\etalchar{+}}20]{brown2020language}
Tom~B Brown, Benjamin Mann, Nick Ryder, Melanie Subbiah, Jared Kaplan, Prafulla
  Dhariwal, Arvind Neelakantan, Pranav Shyam, Girish Sastry, Amanda Askell,
  et~al.
\newblock Language models are few-shot learners.
\newblock {\em arXiv preprint arXiv:2005.14165}, 2020.

\bibitem[BST{\etalchar{+}}18]{benedetto2018creation}
Gary Benedetto, Jordan~C Stanley, Evan Totty, et~al.
\newblock The creation and use of the sipp synthetic beta v7. 0.
\newblock {\em US Census Bureau}, 2018.

\bibitem[CBM{\etalchar{+}}17]{choi2017generating}
Edward Choi, Siddharth Biswal, Bradley Malin, Jon Duke, Walter~F Stewart, and
  Jimeng Sun.
\newblock Generating multi-label discrete patient records using generative
  adversarial networks.
\newblock In {\em Machine learning for healthcare conference}, pages 286--305.
  PMLR, 2017.

\bibitem[CCB15]{cho2015describing}
Kyunghyun Cho, Aaron Courville, and Yoshua Bengio.
\newblock Describing multimedia content using attention-based encoder-decoder
  networks.
\newblock {\em IEEE Transactions on Multimedia}, 17(11):1875--1886, 2015.

\bibitem[CGRS19]{child2019generating}
Rewon Child, Scott Gray, Alec Radford, and Ilya Sutskever.
\newblock Generating long sequences with sparse transformers.
\newblock {\em arXiv preprint arXiv:1904.10509}, 2019.

\bibitem[DBK{\etalchar{+}}20]{dosovitskiy2020image}
Alexey Dosovitskiy, Lucas Beyer, Alexander Kolesnikov, Dirk Weissenborn,
  Xiaohua Zhai, Thomas Unterthiner, Mostafa Dehghani, Matthias Minderer, Georg
  Heigold, Sylvain Gelly, et~al.
\newblock An image is worth 16x16 words: Transformers for image recognition at
  scale.
\newblock {\em arXiv preprint arXiv:2010.11929}, 2020.

\bibitem[DCLT18]{devlin2018bert}
Jacob Devlin, Ming-Wei Chang, Kenton Lee, and Kristina Toutanova.
\newblock Bert: Pre-training of deep bidirectional transformers for language
  understanding.
\newblock {\em arXiv preprint arXiv:1810.04805}, 2018.

\bibitem[DMNS06]{dwork2006calibrating}
Cynthia Dwork, Frank McSherry, Kobbi Nissim, and Adam Smith.
\newblock Calibrating noise to sensitivity in private data analysis.
\newblock In {\em Theory of cryptography conference}, pages 265--284. Springer,
  2006.

\bibitem[DR{\etalchar{+}}14]{dwork2014algorithmic}
Cynthia Dwork, Aaron Roth, et~al.
\newblock The algorithmic foundations of differential privacy.
\newblock {\em Foundations and Trends in Theoretical Computer Science},
  9(3-4):211--407, 2014.

\bibitem[GPAM{\etalchar{+}}14]{goodfellow2014generative}
Ian~J Goodfellow, Jean Pouget-Abadie, Mehdi Mirza, Bing Xu, David Warde-Farley,
  Sherjil Ozair, Aaron Courville, and Yoshua Bengio.
\newblock Generative adversarial nets.
\newblock {\em stat}, 1050:10, 2014.

\bibitem[HVU{\etalchar{+}}18]{huang2018music}
Cheng-Zhi~Anna Huang, Ashish Vaswani, Jakob Uszkoreit, Noam Shazeer, Ian Simon,
  Curtis Hawthorne, Andrew~M Dai, Matthew~D Hoffman, Monica Dinculescu, and
  Douglas Eck.
\newblock Music transformer.
\newblock {\em arXiv preprint arXiv:1809.04281}, 2018.

\bibitem[JYVDS18]{jordon2018pate}
James Jordon, Jinsung Yoon, and Mihaela Van Der~Schaar.
\newblock Pate-gan: Generating synthetic data with differential privacy
  guarantees.
\newblock In {\em International Conference on Learning Representations}, 2018.

\bibitem[KRMW14]{kingma2014semi}
Diederik~P Kingma, Danilo~J Rezende, Shakir Mohamed, and Max Welling.
\newblock Semi-supervised learning with deep generative models.
\newblock {\em arXiv preprint arXiv:1406.5298}, 2014.

\bibitem[PMG{\etalchar{+}}18]{park2018data}
Noseong Park, Mahmoud Mohammadi, Kshitij Gorde, Sushil Jajodia, Hongkyu Park,
  and Youngmin Kim.
\newblock Data synthesis based on generative adversarial networks.
\newblock {\em Proc. VLDB Endow.}, page 1071–1083, 2018.

\bibitem[PVU{\etalchar{+}}18]{parmar2018image}
Niki Parmar, Ashish Vaswani, Jakob Uszkoreit, Lukasz Kaiser, Noam Shazeer,
  Alexander Ku, and Dustin Tran.
\newblock Image transformer.
\newblock In {\em International Conference on Machine Learning}, pages
  4055--4064. PMLR, 2018.

\bibitem[RWC{\etalchar{+}}19]{radford2019language}
Alec Radford, Jeffrey Wu, Rewon Child, David Luan, Dario Amodei, and Ilya
  Sutskever.
\newblock Language models are unsupervised multitask learners.
\newblock {\em OpenAI blog}, 1(8):9, 2019.

\bibitem[SDV20]{sutter2020multimodal}
Thomas~M Sutter, Imant Daunhawer, and Julia~E Vogt.
\newblock Multimodal generative learning utilizing jensen-shannon-divergence.
\newblock {\em arXiv preprint arXiv:2006.08242}, 2020.

\bibitem[SNPT19]{shi2019variational}
Yuge Shi, Siddharth Narayanaswamy, Brooks Paige, and Philip H.~S. Torr.
\newblock Variational mixture-of-experts autoencoders for multi-modal deep
  generative models.
\newblock In {\em NIPS}, 2019.

\bibitem[SS14]{srivastava2014multimodal}
Nitish Srivastava and Ruslan Salakhutdinov.
\newblock Multimodal learning with deep boltzmann machines.
\newblock {\em J. Mach. Learn. Res.}, 15(1):2949--2980, 2014.

\bibitem[UCG{\etalchar{+}}16]{uria2016neural}
Benigno Uria, Marc-Alexandre C{\^o}t{\'e}, Karol Gregor, Iain Murray, and Hugo
  Larochelle.
\newblock Neural autoregressive distribution estimation.
\newblock {\em The Journal of Machine Learning Research}, 17(1):7184--7220,
  2016.

\bibitem[VMP20]{vasco2020mhvae}
Miguel Vasco, Francisco~S Melo, and Ana Paiva.
\newblock Mhvae: a human-inspired deep hierarchical generative model for
  multimodal representation learning.
\newblock {\em arXiv preprint arXiv:2006.02991}, 2020.

\bibitem[VSP{\etalchar{+}}17]{vaswani2017attention}
Ashish Vaswani, Noam Shazeer, Niki Parmar, Jakob Uszkoreit, Llion Jones,
  Aidan~N Gomez, Lukasz Kaiser, and Illia Polosukhin.
\newblock Attention is all you need.
\newblock In {\em NIPS}, 2017.

\bibitem[WG18]{wu2018multimodal}
Mike Wu and Noah Goodman.
\newblock Multimodal generative models for scalable weakly-supervised learning.
\newblock {\em arXiv preprint arXiv:1802.05335}, 2018.

\bibitem[XSCIV19]{xu2019modeling}
Lei Xu, Maria Skoularidou, Alfredo Cuesta-Infante, and Kalyan Veeramachaneni.
\newblock Modeling tabular data using conditional gan.
\newblock In H.~Wallach, H.~Larochelle, A.~Beygelzimer, F.~d\textquotesingle
  Alch\'{e}-Buc, E.~Fox, and R.~Garnett, editors, {\em Advances in Neural
  Information Processing Systems}, volume~32, pages 7335--7345. Curran
  Associates, Inc., 2019.

\end{thebibliography}

% \clearpage
% \appendix
% \section{Appendices}

% \subsection{SDGym results} \label{sdgym_results}

% \subsubsection{Gaussian Mixture datasets}
% \resultstable{ring}
% \resultstable{grid}
% \resultstable{gridr}
% \clearpage

% \subsubsection{Bayesian Networks datasets}
% \resultstable{asia}
% \resultstable{alarm}
% \resultstable{child}
% \resultstable{insurance}
% \clearpage

% \subsubsection{Real World datasets}
% \resultstable{adult}
% \resultstable{census}
% \resultstable{covtype}
% \resultstable{credit}
% \resultstable{intrusion}
% \resultstable{news}
% \clearpage

% \subsection{CelebA results} \label{celeba_results}

% \gansamples{1}{Male, Beard. The top images are generated using the same vector generated using the subset of attributes Male and Beard. Below, one image is fed back into the model and the non-specified attributes are inferred. We give the predicted attribute together with the confidence level.}
% \gansamples{2}{Male, Black Hair, Smiling}
% \gansamples{3}{Female, Blond, Not Smiling}
% \gansamples{4}{Grey Hair}
% \gansamples{5}{Male, Young, Brown Hair, Bangs, Mustache, No Goatee, Mouth slightly open, Bags Under Eyes}
% \clearpage

\end{document}